%% file: main.tex
\documentclass[10pt,twocolumn,letterpaper]{article}

\usepackage{cvpr}              

\input{preamble}

\definecolor{cvprblue}{rgb}{0.21,0.49,0.74}
\usepackage[pagebackref,breaklinks,colorlinks,citecolor=cvprblue]{hyperref}

\usepackage{graphicx}
\usepackage[table]{xcolor}
\usepackage{colortbl}

\usepackage{tikz}
\usetikzlibrary{positioning,fit,calc,spy}

\def\paperID{59} 
\def\confName{3DV\xspace}
\def\confYear{2026\xspace}

\definecolor{lightyellow}{RGB}{255, 255, 204}
\title{Splatography: Sparse multi-view dynamic Gaussian Splatting for filmmaking challenges}

\author{Adrian Azzarelli\\
Bristol Visual Institute\\
University of Bristol, UK\\
{\tt\small a.azzarelli@bristol.ac.uk}
\and
Nantheera Anantrasirichai\\
Bristol Visual Institute\\
University of Bristol, UK\\
{\tt\small n.anantrasirichai@bristol.ac.uk}
\and
David R Bull\\
Bristol Visual Institute\\
University of Bristol, UK\\
{\tt\small dave.bull@bristol.ac.uk}
}

\begin{document}
\twocolumn[{%
\renewcommand\twocolumn[1][]{#1}%
\maketitle
\begin{center}
    \centering
    \captionsetup{type=figure}
    \includegraphics[width=\linewidth]{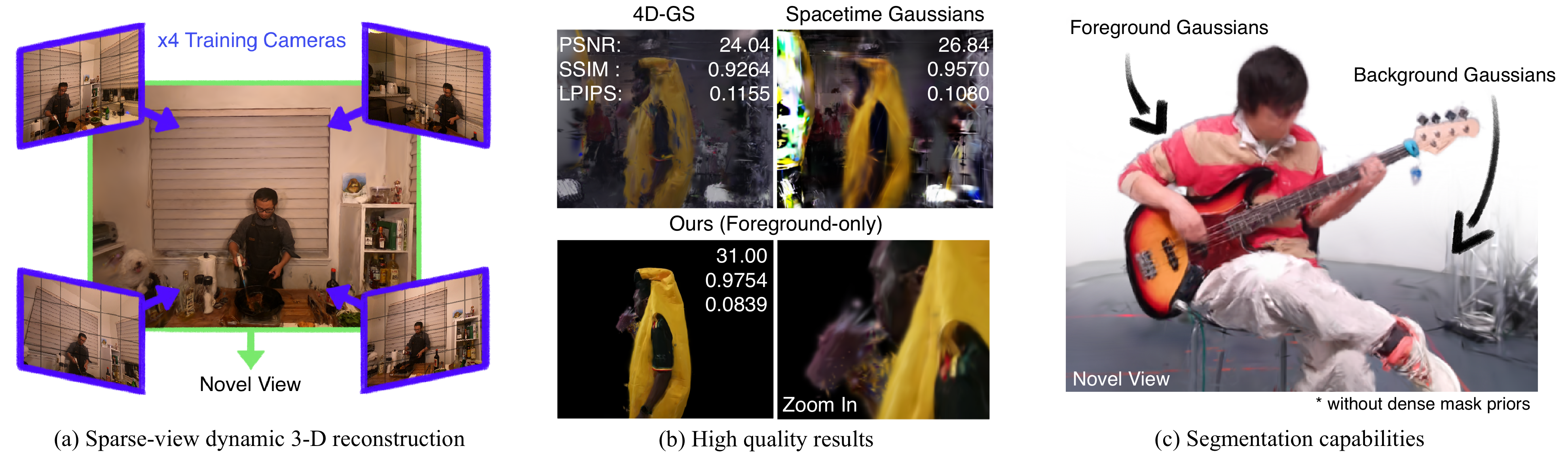}
    \captionof{figure}{\textbf{Sparse view 3-D reconstruction:} Our dynamic representation offers foreground-background separability and high quality 3-D reconstruction without the need for dense mask priors. This paper focuses on filmmaking challenges, including but not limited to sparse view and reflective, transparent and dynamic textures}
\label{fig: poster}
\end{center}%
}]

\maketitle
\begin{abstract}
Deformable Gaussian Splatting (GS) accomplishes photorealistic dynamic 3-D reconstruction from dense multi-view video (MVV) by learning to deform a canonical GS representation. However, in filmmaking, tight budgets can result in sparse camera configurations, which limits state-of-the-art (SotA) methods when capturing complex dynamic features. To address this issue, we introduce an approach that splits the canonical Gaussians and deformation field into foreground and background components using a sparse set of masks for frames at $t=0$. Each representation is separately trained on different loss functions during canonical pre-training. Then, during dynamic training, different parameters are modeled for each deformation field following common filmmaking practices. The foreground stage contains diverse dynamic features so changes in color, position and rotation are learned. While, the background containing film-crew and equipment, is typically dimmer and less dynamic so only changes in point position are learned.
Experiments on 3-D and 2.5-D entertainment datasets show that our method produces SotA qualitative and quantitative results; up to $3$ PSNR higher with half the model size on 3-D scenes. Unlike the SotA and without the need for dense mask supervision, our method also produces segmented dynamic reconstructions including transparent and dynamic textures. Code and video comparisons are available online: 
\url{http://bit.ly/4oqzZrO}
\end{abstract}

\section{Introduction}\label{sec: introduction}
Gaussian Splatting (GS) is increasingly used for photorealistic dynamic 3-D reconstruction from multi-view video (MVV) data. While GS methods can unlock creative potential in filmmaking applications \cite{chen2024gaussianeditor, wu2024gaussctrl, guo2024prtgs, gao2024relightable, huang2024sc, azzarelli2025intelligent}, current approaches \cite{li2024spacetime, wu20244d, huang2024sc, duan20244d, li2024st} rely on a large number of cameras for robust dynamic reconstruction. This can be impractical for some filmmaking productions, where the number of cameras is limited by filming budgets, or issues with acting challenges arise due to isolating filming spaces designed to minimize background noise \cite{cheng2023dna, icsik2023humanrf, qian2024gaussianavatars, li2024animatable}. Hence, improving sparse view 3-D reconstruction (SV3D) for dynamic scenes is a key challenge.

\begin{figure}
    \centering
    \includegraphics[width=1.0\linewidth]{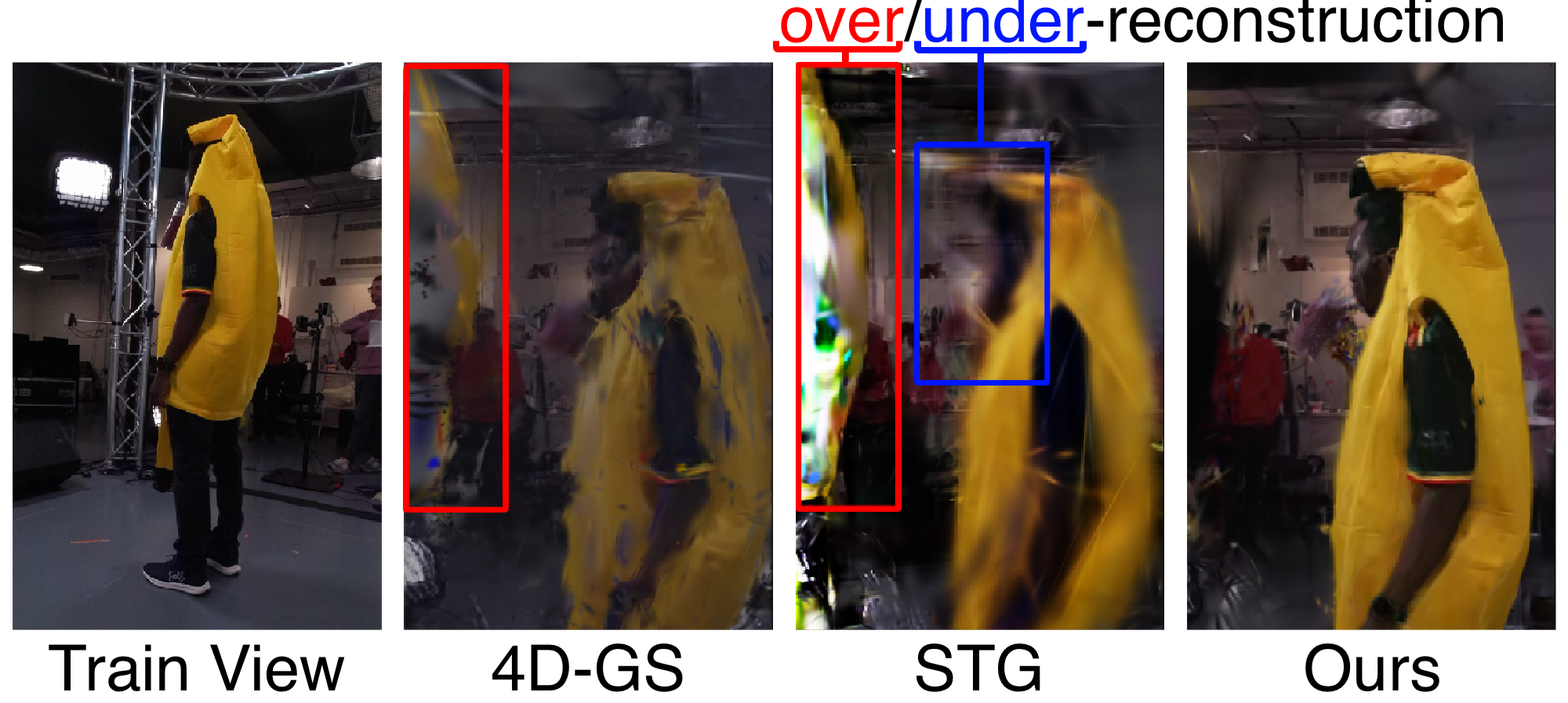}
    \caption{Novel views (right) reveal over-reconstructed backgrounds and under-reconstructed foregrounds in SotA}
    \label{fig: under-over-reconstruction}
\end{figure}

In sparse MVV configurations, fewer cameras typically result in densely viewed foregrounds but sparsely viewed backgrounds. This conflicts with the workings of state-of-the-art (SotA) approaches that attribute Gaussian point importance based on full-image reconstruction quality \cite{wu20244d, li2024spacetime, hu2025learnable}. Thus, backgrounds receive more attention during training due to their larger area, leading to over-reconstruction in backgrounds and under-reconstruction in foregrounds; see \cref{fig: under-over-reconstruction}.
While this could be resolved by masking out backgrounds for each frame using priors generated with SAM2/MiVOS/etc. \cite{ravi2024sam, cheng2021modular}, the issue of densely masking reflective, transparent and dynamic (RTD) textures (like fire and smoke) has not been resolved. Transparency is especially problematic as disentangling dimming and lighting material effects from a background cannot be achieved with simple binary masks. RTD textures are common in filmmaking, so dense mask priors would limit the use of costumes and props. For SV3D we also find that, even before dynamic training, SotA methods cannot reconstruct robust canonical representations. These methods work by learning to deform a canonical (time-independent) 3-D GS. Despite its importance, robust canonical pre-training has yet to be investigated for enhancing dynamic reconstruction. 

Consequently, this paper's primary focus is to tackle SV3D reconstruction challenges by: (1) proposing a new dynamic scene representation that disentangles the foreground and background to deal with point-importance, and (2) developing a new strategy for training canonical representations to deal with poor initialization. Linking this work to filmmaking, we leverage foreground-background separation to apply background-specific constraints based on filmmaking practices; discussed in the following paragraphs. This aims to suppress the reconstruction artifacts associated with background and foreground reconstruction, which we identify in \cref{fig: under-over-reconstruction}. We also place emphasis on reconstructing RTD textures as they are more common in filmmaking than most other practices.

\begin{figure*}
    \centering
    \includegraphics[width=\textwidth]{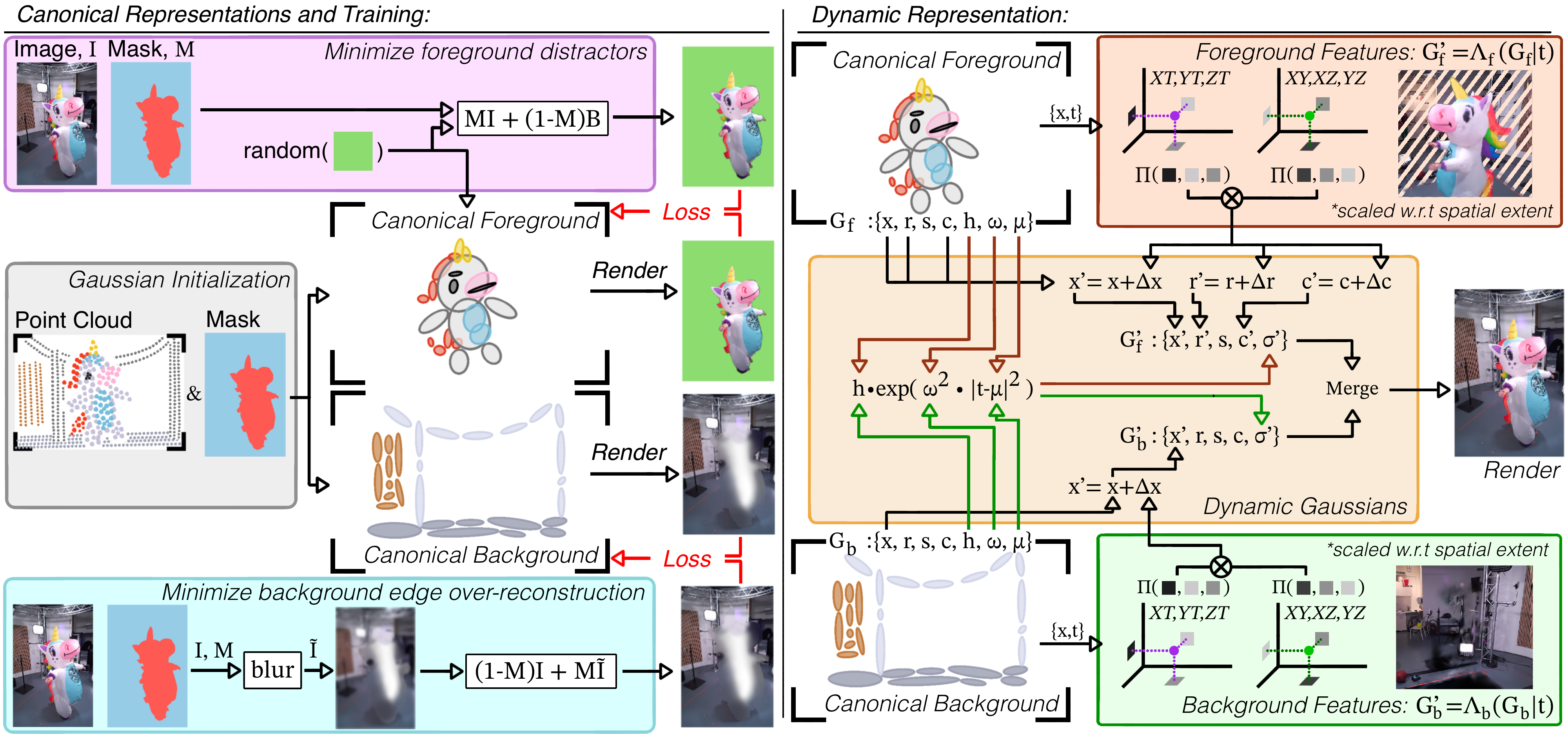}
    \caption{\textbf{Left:} The canonical representation is constructed by masking the initial point cloud and training the foreground and background representations $G_f$ and $G_b$ on specialized loss functions that minimize over-reconstruction. \textbf{Right:} Dynamic features for $G_f$ {and} $G_b$ are jointly trained using the proposed plane-based design. For $G_b$, we only learn motion. For $G_f$, we learn motion, rotation and color change using a novel combination of plane features, and also temporal opacity using an exponential peaking function.}
    \label{fig: main figure}
\end{figure*}

The proposed scene representation in \cref{fig: main figure} uses only a single mask per view at $t=0$ to segment the initial point cloud into canonical foreground and background representations. During canonical training, the representations are separately trained using specialized loss functions that rely on the aforementioned masks. The foreground loss discourages floaters and smooth-edge artifacts, and the background loss suppresses view-dependent occlusions and foreground artifacts present along the mask-edge. To model dynamics, we propose representing the foreground and background as separate hex-plane deformation fields, jointly trained without mask supervision. The assumption made is that, during filming, backgrounds are weakly dynamic as the only temporal changes are due to film crew moving around. Color changes are not expected as background lighting is usually static and dim. So, dynamic background features are captured using only point displacement. By doing this, the background is less likely to over-compensate for under-reconstructed foregrounds. Regarding foreground dynamics, we model changes in position, rotation and color via a hex-plane model as we expect the presence of RTD textures. We also modify the temporal opacity function in \cite{li2024spacetime} to better represent deformable textures and introduce regularizers for each relative parameter.

Dealing with point densification, prior works rely on cloning and splitting strategies that assess volumetric importance (i.e. where to instance points) based on full image reconstruction quality \cite{kerbl3Dgaussians, wu20244d, li2024spacetime}. For SV3D, sparsely viewed backgrounds dominate these metrics leading to over-reconstruction in unimportant regions of space. This also leads to temporal bias, as instances in time with greater reconstruction error may produce clones that negatively affect temporal consistency in other views. We consequently leverage the separation of foreground and background representations to limit cloning only for  foreground points. To avoid view dependent and temporal bias during densification, we propose a reference-free strategy that samples point motion over time and uses quantile selection to clone the most dynamic points. As this approach is reference-free, it does not require tracking additional point parameters for densification, as in prior works.

In summary, this paper aims to resolve current issues when reconstructing dynamic multi-view SV3D content, including RTD textures, via the following contributions:
\begin{enumerate}
    \item A novel approach to disentangling and separately training canonical foreground and background representations using a sparse set of mask; including the proposal of representation-specific loss functions
    \item A new formulation for dynamic representations that treat foregrounds and backgrounds differently by limiting the expressiveness of background features. By also limiting the number of background Gaussians being modeled, this encourages the model to focus more on the foreground representation 
    \item A reference-free and minimally invasive approach to foreground-only point densification. This mitigates issues with current approaches, that utilize full-reference reconstruction quality to mediate point densification
    \item As a by-product of the foreground-background separation, our approach is also capable of foreground segmentation, including RTD textures, with minimal supervision
\end{enumerate}
To our knowledge, this approach is the first to separate and condition a scene based on foreground and background observations. This paper is also the first to investigate dynamic SV3D for scene-based reconstructions and the first to investigate improvements for robust canonical initialization.

\section{Related Work}
In this section we discuss related work on dynamic SV3D. A review of static SV3D research is available in the appendix, however as static methods do not account for RTD textures most solutions are not applicable to the dynamic SV3D paradigm.

\paragraph{General Discussion:}
Research on dynamic 3-D reconstruction began with D-NeRF \cite{pumarola2021dnerf}, an implicit deformable neural radiance field (NeRF). However, it was soon found that key-frame base solutions produced better results \cite{li2022neural,attal2023hyperreel, song2023nerfplayer, wang2023neural, park2023temporal}. DyNeRF \cite{li2022neural} proposed a training strategy for key-frame NeRFs that relied on attributing higher loss (via an importance weight) to pixels with larger motion vectors. HyperReel \cite{attal2023hyperreel} proposed a hierarchical interpolation strategy for a set of tri-plane key-frame representations \cite{chen2022tensorf}. Similarly, HexPlanes \cite{cao2023hexplane} and K-Planes \cite{fridovich2023k} proposed modeling key-frames as 2-D spatiotemporal feature grids, mapping $(x,t), (y,t), (z,t) \rightarrow f_{XT},f_{YT},f_{ZT}$ feature space. Combining this with a tri-plane representation, $(x,y), (x,z), (y,z) \rightarrow f_{XY},f_{XZ},f_{YZ}$, produces a hex-plane feature representation.

More recently, research on dynamic GS has become prevalent \cite{wu20244d, li2024spacetime, huang2024sc, guo2024motion, yang2023real, hu2025learnable}.
Since, we have noticed a divide between fully explicit and implicit dynamic representations. Explicit models \cite{li2024spacetime, huang2024sc, yang2023real, gao2024gaussianflow} learn to deform Gaussian points by, for example, using polynomials to model motion \cite{li2024spacetime}. Polynomial functions offer the opportunity to apply parameter-specific constraints, but require more computation than implicit solutions and are degenerate for longer videos. Instead, implicit representations like hex-planes \cite{yan20244d,wu20244d,huang2025adc, azzarelli2023waveplanes} learn features that have a one-to-many relationship w.r.t GS point parameters. They are compact and do not degenerate w.r.t video duration but are harder to precisely constrain dynamic components.

\paragraph{Filmmaking Applications:}
Research on 3-D human-centric/avatar reconstruction has tended to focus on cases where dense mask priors are attainable \cite{icsik2023humanrf, qian2024gaussianavatars, li2024animatable, jung2023deformable, pang2024ash}. However, more recently there has been increased interest in applying these methods to sparse view configurations \cite{jin2025diffuman4d}. Since these methods rely on dense mask priors in order to achieve high quality foreground reconstructions, significant problems remain, particularly when applying them to scenes that capture dynamic and transparent textures.

For scene-based reconstruction, \cite{azzarelli2025vivo} tested 4D-GS \cite{wu20244d}, STG \cite{li2024spacetime} and SCGS \cite{huang2024sc} on several entertainment scenes using a multi-view SV3D set-up. 4D-GS achieves reconstruction using the 3D-GS \cite{kerbl3Dgaussians} model with spatiotemporal changes driven by a K-Planes feature representations. It uses the Gaussian cloning and splitting strategy, proposed in \cite{kerbl3Dgaussians}. STG introduces a representation and rendering method driven by a temporal event-based parameter. The parameter is involved in polynomial point displacement, temporal rotation functions, and in an exponential change in opacity function. The proposed densification strategy optimizes the number of Gaussians in sparsely viewed spatial regions. SC-GS is a hybrid approach for mesh-like Gaussians, where motion is driven by a small set of local control points and the linked Gaussians are trained as surface-like residuals using a rigidity loss to maximize uniform local motions.

These methods place equal importance across the whole scene. Hence, when the spatial extent of the background is much greater than the foreground, the background receives more attention during training and densification. As backgrounds in SV3D are non-salient across view, this leads to under-reconstructed foreground and over-reconstructed background content. This analysis is supported by \cref{fig: under-over-reconstruction} and highlights the need to investigate the dynamic SV3D paradigms. 

\section{Preliminaries}\label{sec: preliminaries}
In \cref{sec: preliminaries.canon} we present background on 3-D GS rendering and in \cref{sec: preliminaries.hex} we present background hex-plane deformation fields, which we use to model the various dynamic foreground and background parameters.

\subsection{Canonical 3-D Gaussians}\label{sec: preliminaries.canon}
3-D Gaussians \cite{kerbl3Dgaussians} are characterized as $G = \{x, s, r, c, \sigma \}$, representing point position $x \in \mathcal{R}^3$, covariance constructed from $\Sigma = RSS^TR^T$, where $R$ is the $3 \times 3$ rotation transform represented by $r \in \mathcal{R}^4$ in quaternion form and $S$ is the scaling transform represented as $s \in \mathcal{R}^3$, color $c \in \mathcal{R}^3$ and opacity $\sigma \in \mathcal{R}$. The set $G$ is differentiably rendered as a 2-D image via splatting \cite{zwicker2002ewa}. This involves: (1) sorting $G$ w.r.t point distance from the camera, (2) projecting $\Sigma$ into image space using $\Sigma' = JW\Sigma W^TJ^T$, where $W$ is the viewing transform, $J$ is a Jacobian of the affine approximation of the projective transform and $\Sigma'$ is the 2-D covariance of Gaussian on the image plane, and (3) alpha blending 2-D splats using the formula for pixel color:
\begin{equation}
    C = \sum_{i \in N} c_i \alpha_i \prod_{j=1}^{i-1}(1-\alpha_i),
\end{equation}
where $i$ is the index of each splat's color and alpha w.r.t the aforementioned sorting and $\alpha$ is the influence of a splat on a pixel with displacement $\mathcal{X}$ from the 2-D splat's center, given by the Mahalanobis distance function:
\begin{equation}\label{eq: mahalanobis}
    \alpha = \sigma \cdot\text{exp}(-\frac{1}{2} \mathcal{X}^T \Sigma^{-1}\mathcal{X}).
\end{equation}

Ultimately, we choose MipSplatting \cite{yu2024mip} as our canonical representation as it resolves 3D-GS anti-aliasing due to non-salient object resolutions across views. Knowledge of MipSplatting is not required to understand the main paper so details are provided in the appendix.

\subsection{Hex-Plane Deformation Fields}\label{sec: preliminaries.hex}
Hex-Planes \cite{fridovich2023k, wu20244d, cao2023hexplane, azzarelli2023waveplanes, huang2024sc} model the change in each Gaussian parameter over time, $t$. The point position, $x$, is concatenated with the time $t$, forming $q=\{x, y, z, t\}$. The normalized 2-D coordinates $q_{\psi}$ are projected onto a grid-plane $P_{\psi} \in \mathcal{R}^{J\times K\times L}$ for $\psi \in [XY, XZ, YZ, XT, YT, ZT]$ and bilinearly sample a feature $f_{\psi}=\theta (q_{\psi})$, where $J\times K$ is the pixel resolution of the grid-planes, $L$ is the feature size, and $\theta$ represents the projection and sampling operations. Per point, all features are multiplied forming $f(q)=\prod_{i \in \psi}f_i(q)$.  The final feature is decoded via a shallow MLP per Gaussian component. This entire process is defined as $\Lambda(G|t)$ leading to $G' = G + \Lambda(G|t)$, where $G'$ is the deformed representation at time-step $t$.

\section{Methodology}
Our approach to tackling dynamic SV3D for MVV content is summarized in \cref{fig: main figure}. \cref{sec: canonical gaussians} presents our canonical representation and optimization strategy, responsible for decoupling and separately training background and foreground Gaussians. \cref{sec: representation} presents our formulation of the foreground and background deformation fields, and optimization. In \cref{sec: opacity} we present the temporal opacity function and regularization approach, and  in \cref{sec: densification} we propose our reference-free densification strategy, which focuses on reducing under-reconstruction artifacts in the foreground. Further details are available in the appendix.

\subsection{Canonical Foreground and Background Gaussians}\label{sec: canonical gaussians}
\paragraph{Representation.} 
Accurate initialization is critical for dynamic reconstruction and works to optimize the quality of the canonical representation before learning dynamics. Prior work \cite{wu20244d, li2024spacetime, fridovich2023k, azzarelli2023waveplanes} has tended to over-reconstruct backgrounds and under-reconstruct foregrounds during this stage of SV3D reconstruction. To address this imbalance, we first separate the initial point cloud into foreground and background GS representations, $G_f$ and $G_b$ respectively, using a sparse set of masks, $M^* \in \mathcal{R}^{H\times W}$ for MVV frames at $t=0$. This can be accomplished by projecting the initial point cloud onto the 2-D plane of each mask and assigning points that appear in all masks to $G_f$. 

Considering that our approach to densification is fully explicit (see in \cref{sec: densification}), we can estimate the final number of points with reasonable accuracy. This facilitates  mitigation of over and under-reconstruction errors that arise from $G_f$ and $G_b$ being too sparsely or densely populated at the beginning of training. This is done by applying voxel up/down sampling to the initial $G_f$ and $G_b$ representations. In all of our experiments (on 1080p and 2K resolution images) we found $\sim 50,000$ points to be sufficient.

\paragraph{Optimization.}
Prior works that rely on hex-planes to model deformations \cite{wu20244d, azzarelli2023waveplanes, huang2024sc} train $G$ during the canonical training stage using all the ground truth images, $I^* \in \mathcal{R}^{H\times W\times 3}$ for $t \ge 0$. This means that $G$ in not interpretable by the renderer without using $G'=\Lambda(G|t)$ to update $G$. This makes developing loss functions for canonical training challenging. Instead, during the this stage we anchor $G_f$ and $G_b$ to $t=0$ by only using ground truth data at $t=0$. This means $G_f$ and $G_b$ can be rendered as stand-alone components. This allows us to develop the following mask-based loss functions to deal with the challenges discussed in \cref{sec: introduction}.

To minimize foreground floaters and smooth-edge artifacts, we propose \cref{eq: foreground blended loss} for training $G_f$, where the first component alpha blends a random background color, $B \in \mathcal{R}^{3}$, with the masked ground truth image and the second component alpha blends $B$ with the image rendered from $G_f$. Here, $\alpha_f \in \mathcal{R}^{H\times W}$ is the alpha of $G_f$. $B$ is randomly sampled every iteration. 
\begin{equation}\label{eq: foreground blended loss}
    \mathcal{L_\text{f}} = ||(M^{*}I^* + (1-M^{*})B) - (\alpha_{f}I_f + (1-\alpha_f)B)||
\end{equation}

To avoid over-fitting background colors that fill in the gaps left by under-reconstructed foreground volumes, we optimize the similarity between the edge Gaussians and surrounding Gaussians using \cref{eq: background edge loss}. As in \cref{fig: main figure}, this bleeds the edges of the background into the masked region using $\tilde{I}^b = \text{blur}((1-M^*)I^{*})$ and alpha blends the blurred and original image to preserve the background detail. 
\begin{equation}\label{eq: background edge loss}
    \mathcal{L_\text{b}} = ||((1-M^*)I^{*} + M^*\tilde{I}^{b}) - I^b||
\end{equation}

\begin{table*}[ht!]
    \centering
    \begin{tabular}{c|c c | c c | c c | c c | c }
        Method & \multicolumn{2}{c}{PSNR} & \multicolumn{2}{c}{SSIM} & \multicolumn{2}{c}{LPIPS-Alex}  & \multicolumn{2}{c|}{LPIPS-VGG} &  Size\\
         & Full & Mask & Full & Mask & Full & Mask & Full & Mask & (MB)\\ \hline \hline

         \multicolumn{10}{c}{3-D ViVo dataset \cite{azzarelli2025vivo}: Bassist, Pianist, Weights, Fruit, Pony} \\ \hline
 4D-GS \cite{wu20244d} & 14.22 & 21.22 & 0.6727 & 0.9328 & 0.3832 & 0.1160 &0.4281	& 0.1449& 134\\
 STG \cite{li2024spacetime} & 13.83	&21.72	&0.6759	&0.9398	& 0.4005	&0.1189&0.4276	&0.1431	&280\\
 SC-GS \cite{huang2024sc} & 13.81	&20.57	&0.4244	&0.8536	&0.5047 &0.1727&0.5320	&0.1906 & 320\\ \hline
 Ours & \cellcolor{lightyellow}16.05 &\cellcolor{lightyellow}24.80&\cellcolor{lightyellow}0.7503 & \cellcolor{lightyellow}0.9537 &\cellcolor{lightyellow} 0.3245 & \cellcolor{lightyellow}0.0792 & \cellcolor{lightyellow}0.3839 &\cellcolor{lightyellow} 0.1270 & 60\\ \hline \hline
         
         \multicolumn{10}{c}{2.5-D DyNeRF dataset \cite{li2022neural}:  Flame steak, Flame salmon, Cook spinach}\\ \hline
STG	    &20.81	&21.71	&0.8456	&0.9851	&0.1523	&0.0127	&0.2499	&0.0157 & 34\\
ITGS \cite{hu2025learnable}	&21.95	&24.93	&0.8312	&0.9867	&0.1908	&0.0114	&0.2824	&0.0154 & 119 \\
4D-GS	&\cellcolor{lightyellow}24.51 & 26.45& 0.8548 & 0.9871 &\cellcolor{lightyellow} 0.1308 & 0.0118 &\cellcolor{lightyellow} 0.2265 & 0.0155 &  37\\
W4D-GS \cite{azzarelli2023waveplanes}&24.32	&\cellcolor{lightyellow}26.56	&0.8552	&\cellcolor{lightyellow}0.9873	&0.1319	&0.0117	&0.2294	&0.0157 & 46\\\hline
Ours &24.41 & 26.28& \cellcolor{lightyellow}0.8606 & 0.9872 & 0.1444 &\cellcolor{lightyellow} 0.0110 & 0.2493 & \cellcolor{lightyellow}0.0149 & 47 \\
    \end{tabular}
    \caption{Average results from the 2.5-D and 3-D entertainment datasets. The selected scenes are named alongside each dataset. \textit{Full/Mask} evaluates full and foreground-masked test videos.}
    \label{tab: average results}
\end{table*}

\subsection{Foreground and Background Deformation Fields}\label{sec: representation}
\paragraph{Representation.} 
4D-GS \cite{wu20244d} introduces the hex-plane deformation model $G'=\Lambda(G|t)$ to approximate the linear deformations $(\Delta x, \Delta r, \Delta s, \Delta c, \Delta \sigma)$ for all components in $G$ w.r.t time. This is also applicable to $G_f$ and $G_b$. Though, as $G_f$ and $G_b$ contribute differently to the final render, we choose to formulate $G'_f$ and $G'_b$ differently. The foreground model $G'_f=\Lambda_f(G_f|t)$ is hex-plane representation that approximates $(\Delta x, \Delta r, \Delta c) \forall \in G_f$. The background model $G'_b = \Lambda_b(G_b|t)$, is a different hex-plane representation that approximates $(\Delta x) \forall \in G_b$. $\Lambda_f(\cdot)$ and $\Lambda_b(\cdot)$ are scaled w.r.t the spatial extents of the initial $G_f$ and $G_b$ point clouds, respectively.   

The expressivity of $\Lambda_b$ is limited to $\Delta x$ following assumptions made regarding common filmmaking practices. The assumptions are that the area around the performance stage (the background) is dimly lit, and the film crew limit movement to avoid distracting the performer. As dim background lighting reduces the likelihood of dynamic texture effects, learning $(\Delta r, \Delta c)$ for $G'_b$ imposes an unnecessary challenge for reconstruction. Conversely, foregrounds are assumed to contain a broad range of texture effects, so they learn the full set of dynamic parameters, except $\Delta s$ following \cite{li2024spacetime}.   

\paragraph{Optimization.}
We train dynamic content by merging $(G_b', G_f')$ and rendering the image $I_t$ for timestep $0\leq t$ and applying the panoptic loss,
\begin{equation}
    \mathcal{L} = ||I_t^* - I_t|| \quad \text{for} \quad 0 \le t \le 1
\end{equation}

\subsection{Dynamic Textures via Opacity Constraints}\label{sec: opacity}
\paragraph{Representation.}
Event-based opacity \cite{li2024spacetime} is effective at reconstructing RTD textures. This models $\sigma(t)$ as a normal distribution, where the half-life, mean temporal position and amplitude are learnable components of $G$. Thus, dynamic textures like fire have a short life and solid materials have long life. To better represent these opposing features, we modify the original function by replacing $\omega$ from \cite{li2024spacetime} with $\omega^2$ to bias learning either instantaneous or stationary $0 \le \sigma(t) \le 1$.
\begin{equation}
\sigma(t) = h \cdot \text{exp}(-\omega^2 |t-\mu|^2),
\end{equation}
where $h$ is the peak opacity, $\omega$ is the bandwidth and $\mu$ is the temporal center. This evolves $G_f$ and $G_b$ to include $\{x,r,s,h,\omega, \mu, c \}$.

\paragraph{Optimization.} For training we propose regularizing $h$ and $\omega$ using
\begin{equation}
    \mathcal{L}_{h,\omega} = \lambda_h|1-h| + \lambda_{\omega} |\omega|,
\end{equation}
where $\lambda_h=0.1$ and $\lambda_{\omega}=1.$. This encourages points to be dense and temporally consistent.

\begin{figure*}[ht!]
    \includegraphics[width=\linewidth]{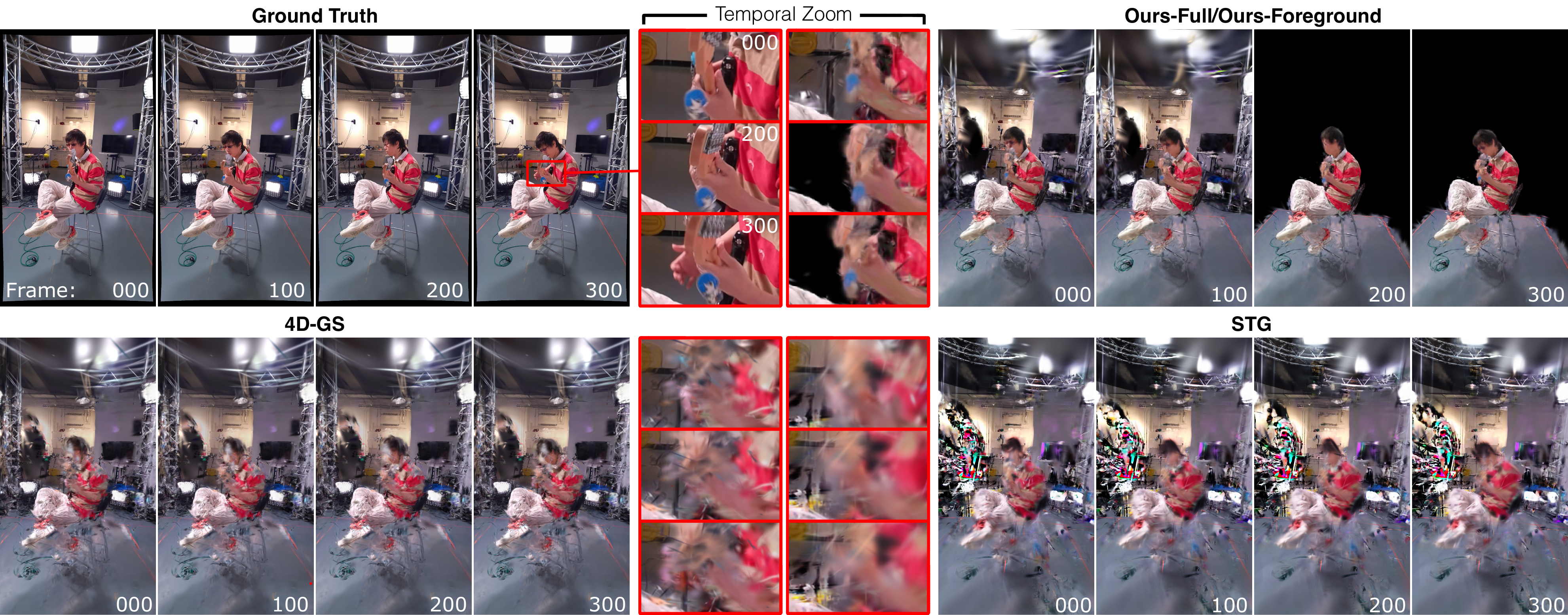}
    \caption{\textbf{Full and Zoom Temporal Comparison:} The zoom results show that our method is the only one capable of capturing the visual dynamics of the semi-transparent key-chain. Using the ViVo-Bassist scene \cite{azzarelli2025vivo}}
    \label{fig: vivo main results}
\end{figure*}

\subsection{Densification}\label{sec: densification}
Prior works on adaptive density for dynamic GS \cite{kerbl3Dgaussians, wu20244d} track parameters that indicate point importance based on full-reference reconstruction quality. In SV3D, sparsely viewed background dominate the full-image metrics leading to higher point instancing in less important background regions, leaving the foreground under-reconstructed. To handle over-reconstructed backgrounds, we initialize $G_b$ with sufficient points to avoid requiring densification during training; see \cref{sec: canonical gaussians}. In contrast, under-reconstruction in foregrounds predominantly occurs in regions exhibiting larger point displacement with few initial points. To address this, we densify points with higher temporal displacement statistics. To identify these points, we linear sample $x' \in G'_f(x, t)$ across 10 time-steps, from $t=0$ to $t=1$. We then find the displacement of each point position $x'_t$, w.r.t to the mean position over time, $\tilde{x}'$, using the Euclidean norm $X'_t = ||x'_t - \tilde{x}'||_2$. The average change in displacement is then computed with $\tilde{X}' = \frac{1}{10} \sum_{t=0}^{10} X'_t$, and thresholded via quantile selection, where the top 10\% largest $\tilde{X}'$ value are selected for point duplication. 

An estimate of the final foreground point count can be made with $N_{final} = ((2^{n_{c}}\times N_{start})\times 1.1)^{n_{d}}$, where $n_{c}$ and $n_{d}$ are the number of times that we densify during canonical and dynamic training stages, respectively. This provides guidance when initializing $G_f$ and $G_b$, as in \cref{sec: canonical gaussians} though it could be helpful for work that focuses on scene compression \cite{kwan2023hinerv, kwan2024immersive}.

\begin{figure*}[ht]
    \includegraphics[width=\linewidth]{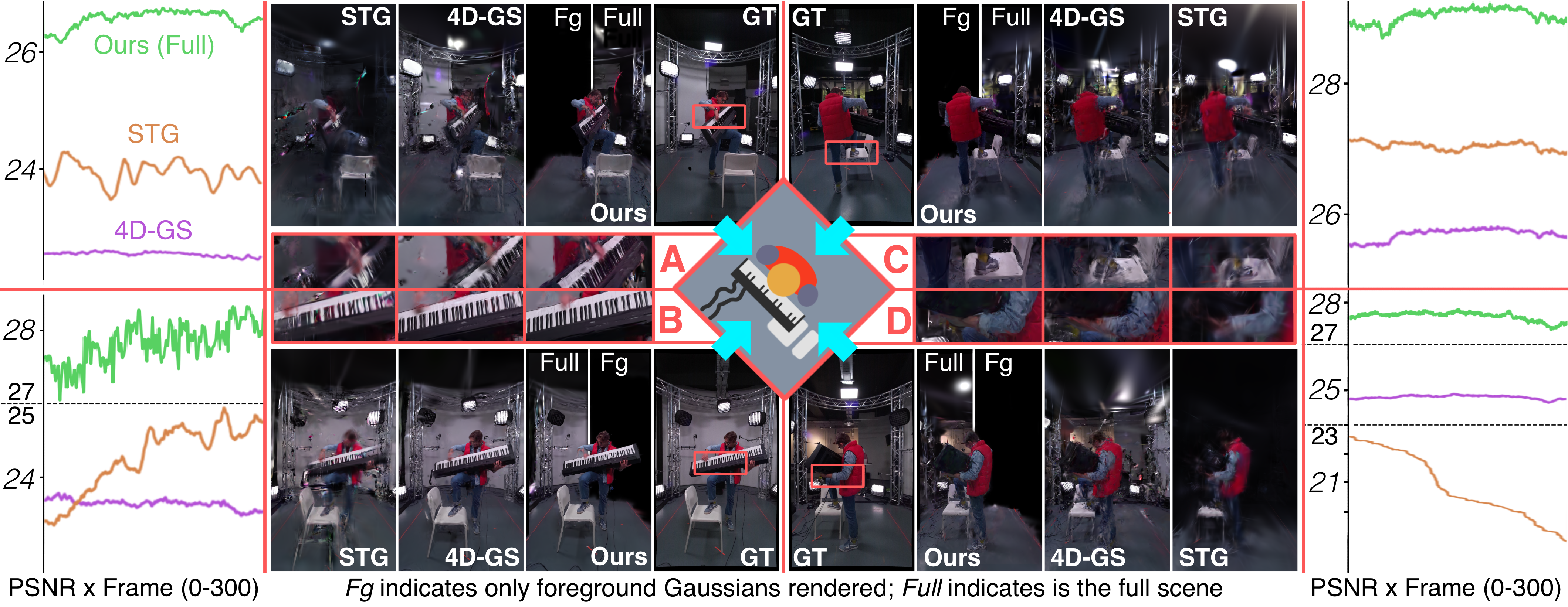}
    \caption{\textbf{Per-Frame and Per-View PSNR Plot:} The surrounding plots show the PSNR result and objectively demonstrate our approach is consistently performant. \textbf{Full and Zoom Frame Comparison:} Our-Foreground (labeled \textit{Ours}) reconstructs keyboard, arms and feet with more visual appeal. Using the ViVo-Pianist scene \cite{azzarelli2025vivo}}
    \label{fig: vivo main results 2}
\end{figure*}
\begin{figure*}[ht]
    \includegraphics[width=\linewidth]{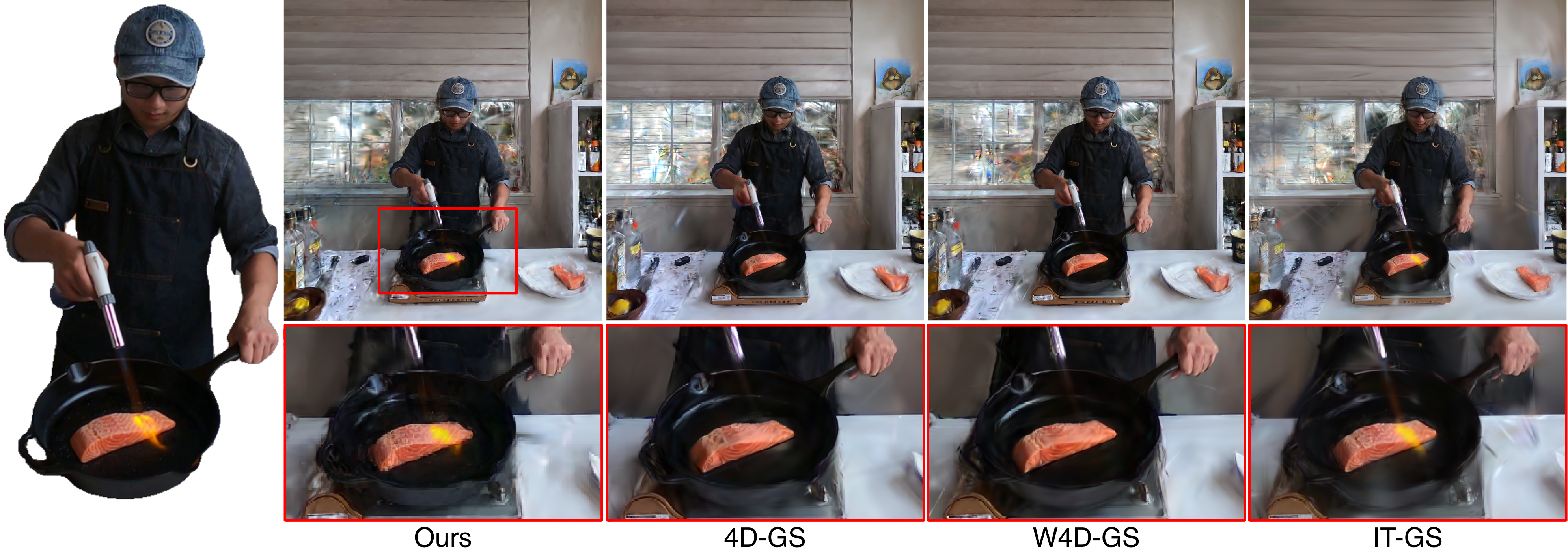}
    \caption{\textbf{Frame-by-frame test view} on the sparse DyNeRF dataset \cite{li2022neural}; 4 training cameras at the extremities were used instead of 20. The masked ground truth is on the left}\label{fig: dynerf results}
\end{figure*}

\section{Experiments}\label{sec: experiments}
This section evaluates performance on two entertainment datasets, \cite{azzarelli2025vivo, li2022neural}. All tests used an Nvidia RTX 3090 (24GB of VRAM). In the appendix and online, we provide more frame-by-frame and video comparisons.

\subsection{Real 3-D Cinematographic Scenes}
We select five scenes from the ViVo dataset \cite{azzarelli2025vivo}. The sequences are trained with 10 cameras for 300 frames with 2K resolution. There are four test cameras placed in-front, behind, left and right of the performance stage. This serves to assess the quality on sparse views that \textit{do not} share background features.

Visual and metric results are provided in \cref{fig: vivo main results,fig: vivo main results 2} and \cref{tab: average results}. Comparatively, the temporal quality in our method is higher and smoother, as shown by the consistency of the per-frame and per-camera PSNR results in \cref{fig: vivo main results 2}. Our method is also capable of producing clean segmentations with only a sparse set of masks, and more accurately captures fine details such as the semi-transparent key-chain in \cref{fig: vivo main results}, which none of the other benchmarks accomplishes.

\begin{figure*}[ht]
\begin{tikzpicture}
    \node[anchor=south west, black] at (-.0, -0.){\includegraphics[width=\linewidth]{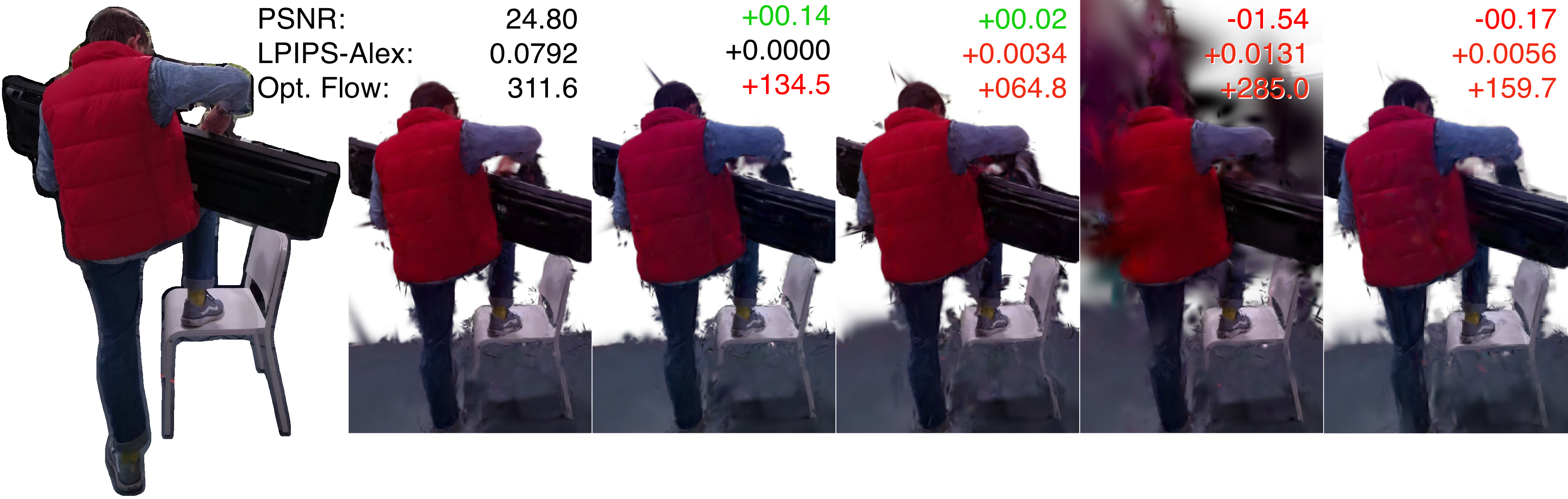}};
    \node[anchor=south west, black] at (4.8, 0.1){\strut Ours};
    
    \node[anchor=south west, black] at (7.3, 0.25){\strut Unified $\Lambda$};
    \node[anchor=south west, black] at (6.75, -0.1){\strut in \cite{wu20244d, azzarelli2023waveplanes, huang2024sc, fridovich2023k}};
    
    \node[anchor=south west, black] at (9.3, 0.25){\strut Canonical Training};
    \node[anchor=south west, black] at (9.8, -0.1){\strut in \cite{wu20244d, azzarelli2023waveplanes, huang2024sc}};

    \node[anchor=south west, black] at (12.5, 0.25){\strut Densification};
    \node[anchor=south west, black] at (12.55, -0.1){\strut in \cite{kerbl3Dgaussians, wu20244d, azzarelli2023waveplanes}};

    \node[anchor=south west, black] at (14.8, 0.25){\strut Temporal Opacity};
    \node[anchor=south west, black] at (15.3, -0.1){\strut in \cite{li2024spacetime, hu2025learnable}};
\end{tikzpicture}
    \caption{\textbf{Visual and average metric results of Ablation.} PSNR and LPIPS-Alex are masked evaluations of the ground truth foreground (left). \textit{Opt. flow} is a MSE optical flow metric (see appendix)}\label{fig: ablations}
\end{figure*}
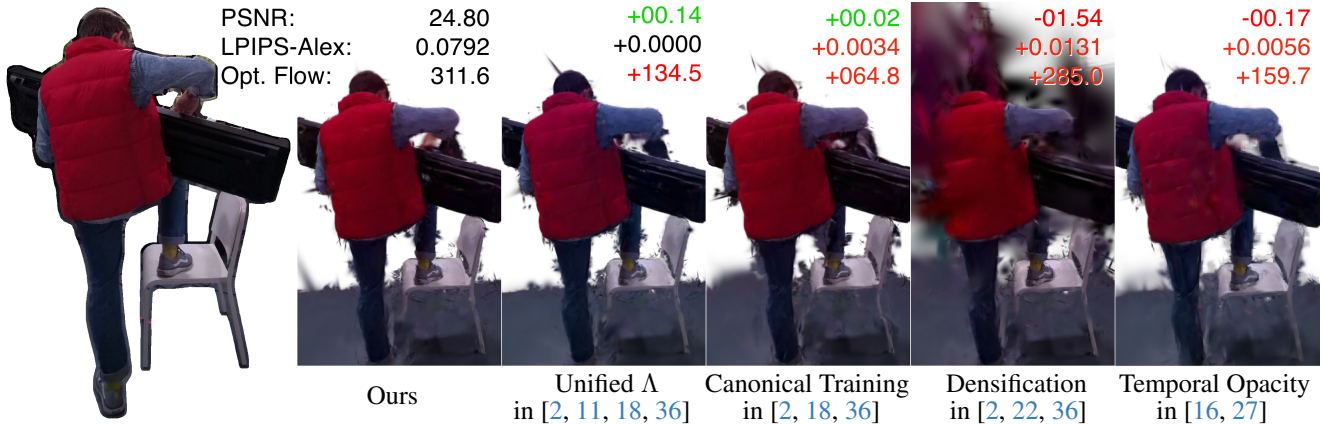

\subsection{Real 2.5-D Cooking Scenes}
We select the three cooking scenes from the DyNeRF dataset that focus on RTD textures. This is an SV2.5D data set and is representative of talk-show productions where all cameras point in the same direction. For training, we select the four most distant cameras from the test camera, for 50 frames at 1080p resolution. This experiment serves to assess the quality on sparse views that \textit{do} share background features.

\cref{tab: average results} and \cref{fig: dynerf results} show the metric and visual results. While 4D-GS and W4D-GS produce competitive results, they fail to reconstruct the fire texture in \cref{fig: dynerf results}. Comparatively, our approach performs well on both visual and metric tests and is capable of reconstructing RTD textures for all scenes. Several scenes also produce high-quality segmentations, demonstrating that our method is capable of producing segmented representations on a variety of camera configurations.

\subsection{Ablations}\label{sec: ablations}
Results of our ablation study are shown  in \cref{fig: ablations}. By unifying $\Lambda_f$ and $\Lambda_b$ such that $G'\rightarrow\Lambda(\{ G_f,G_b\}|t)$ as in \cite{wu20244d, azzarelli2023waveplanes, huang2024sc, fridovich2023k}, the quantitative results indicate little difference but the qualitative and optical flow results reveal weaknesses reconstructing high-frequency temporal details such as hands. Our canonical training strategy is compared to the strategy used in \cite{wu20244d, azzarelli2023waveplanes, huang2024sc}, where $G_f$ and $G_b$ are jointly trained with $\mathcal{L} = ||I^*_t-I_t||$ for $0\leq t \leq 0$. Our method produces better qualitative results, including cleaner masks. Then, using the original densification strategy as in \cite{kerbl3Dgaussians, wu20244d, azzarelli2023waveplanes} is shown to be un-suitable for SV3D problems. Finally, we compare the original temporal opacity function in \cite{li2024spacetime, hu2025learnable}, where $\sigma = h \cdot \text{exp}(\omega |t-\mu|^2)$. The results show our method produces better qualitative and quantitative results, including better high-frequency temporal details, like the hands. 

\section{Limitations}

As with dense-view dynamic GS research \cite{wu20244d, li2024spacetime, huang2024sc, azzarelli2023waveplanes}, issues exist with disentangling view-dependent color changes and point motions in SV3D settings. This occurs due to fast and large motions, as seen by the hand and hand-shadow in \cref{fig: ablations} (Ours), where the shadow blends in with the background; over-reconstructing color and under-reconstructing motion. Following work on human-centric reconstruction \cite{jin2025diffuman4d}, future work could investigate depth-based learning using real/pseudo depth to improve geometric observations. Depth sensors/data are not uncommon in film-making.

\section{Conclusion}
This paper present a new dynamic gaussian splatting pipeline for sparse view 3-D reconstruction from multi-view video, specifically targeting cinematographic applications. This starts by separating the densely viewed foreground and sparsely viewed backgrounds using a sparse set of masks for each frame at $t=0$. Each representation is then separately trained on different loss functions that utilize the sparse masks to minimize over-reconstruction in foreground and backgrounds. The deformation field is also split into foreground and background components and re-formulated to reflect common filmmaking practices;   backgrounds are assumed to be less dynamic and dimmer so only point displacements are modeled for these Gaussians. We further propose a new reference-free Gaussian densification approach, and also modify the temporal opacity function to improve opaque or transparent texture reconstruction.  To evaluate the efficacy of our approach for cinematographic applications, we benchmark our method on sparse view 3-D and 2.5-D datasets that simulate cinematographic content. Not only does our method produce SotA quantitative and qualitative results, especially outperforming the next best method by $>3$ PSNR with half the model size on 3-D scenes, the separation of foreground and background also provides the ability to segment important foreground features - ready for post-production. 

{
    \small
    \bibliographystyle{ieeenat_fullname}
    \bibliography{main}
}

\appendix

\newpage

\end{document}

%% file: preamble.tex
\usepackage[dvipsnames]{xcolor}
